\documentclass[10pt,journal,compsoc]{IEEEtran}
\ifCLASSOPTIONcompsoc
  \usepackage[nocompress]{cite}
\else
  \usepackage{cite}
\fi

\usepackage{graphicx}
\usepackage{amsmath}
\usepackage{amssymb}
\usepackage{array}
\usepackage{enumitem}
\usepackage{stfloats}
\usepackage{booktabs}

\usepackage[pagebackref=true,breaklinks=true,letterpaper=true,colorlinks,bookmarks=false]{hyperref}

\newcommand\etal{\emph{et al}. }

\newcommand{\name}{fully convolutional network}
\newcommand{\Name}{Fully convolutional network}
\newcommand{\NAME}{FCN}
\newcommand{\minisection}[1]{\textbf{#1}\hspace{0.3em}}
\newcommand{\wordsection}[1]{\textbf{#1}}

\newcommand{\metr}[1]{\parbox[b]{0.2in}{#1}}
\newcommand{\pixacc}{\metr{pixel\\acc.}}
\newcommand{\classacc}{\metr{mean\\acc.}}
\newcommand{\meanIU}{\metr{mean\\IU}}
\newcommand{\fwIU}{\metr{f.w.\\IU}}

\begin{document}
\title{Fully Convolutional Networks\\for Semantic Segmentation}
\author{Evan~Shelhamer$^*$\thanks{$\;^*$Authors contributed equally},~Jonathan~Long$^*$,~and~Trevor~Darrell,~\IEEEmembership{Member,~IEEE}%
\IEEEcompsocitemizethanks{\IEEEcompsocthanksitem E. Shelhamer, J. Long, and T. Darrell are with the Department of Electrical Engineering and Computer Science (CS Division), UC Berkeley. E-mail: \{shelhamer,jonlong,trevor\}@cs.berkeley.edu.}
}%

\IEEEtitleabstractindextext{%
\begin{abstract}
Convolutional networks are powerful visual models that yield hierarchies of features.
We show that convolutional networks by themselves, trained end-to-end, pixels-to-pixels, improve on the previous best result in semantic segmentation.
Our key insight is to build ``fully convolutional'' networks that take input of arbitrary size and produce correspondingly-sized output with efficient inference and learning.
We define and detail the space of \name s, explain their application to spatially dense prediction tasks, and draw connections to prior models.
We adapt contemporary classification networks (AlexNet, the VGG net, and GoogLeNet) into \name s and transfer their learned representations by fine-tuning to the segmentation task.
We then define a skip architecture that combines semantic information from a deep, coarse layer with appearance information from a shallow, fine layer to produce accurate and detailed segmentations.
Our \name\ achieves improved segmentation of PASCAL VOC (30\% relative improvement to 67.2\% mean IU on 2012), NYUDv2, SIFT Flow, and PASCAL-Context, while inference takes one tenth of a second for a typical image.
\end{abstract}

}

\maketitle

\IEEEdisplaynontitleabstractindextext
\IEEEpeerreviewmaketitle
\IEEEraisesectionheading{\section{Introduction}\label{sec:introduction}}

\IEEEPARstart{C}{onvolutional} networks are driving advances in recognition.
Convnets are not only improving for whole-image classification \cite{AlexNet, VGGNet, GoogLeNet}, but also making progress on local tasks with structured output.
These include advances in bounding box object detection \cite{OverFeat, RCNN, SPP}, part and keypoint prediction \cite{Ning, Jon}, and local correspondence \cite{Jon, BroxSIFT}.

The natural next step in the progression from coarse to fine inference is to make a prediction at every pixel.
Prior approaches have used convnets for semantic segmentation \cite{ning2005automatic, Ciresan, Farabet, Pinheiro, Bharath, Saurabh, N4}, in which each pixel is labeled with the class of its enclosing object or region, but with shortcomings that this work addresses.

We show that \name s (\NAME s) trained end-to-end, pixels-to-pixels on semantic segmentation exceed the previous best results without further machinery.
To our knowledge, this is the first work to train \NAME s end-to-end (1) for pixelwise prediction and (2) from supervised pre-training.
Fully convolutional versions of existing networks predict dense outputs from arbitrary-sized inputs.
Both learning and inference are performed whole-image-at-a-time by dense feedforward computation and backpropagation.
In-network upsampling layers enable pixelwise prediction and learning in nets with subsampling.

This method is efficient, both asymptotically and absolutely, and precludes the need for the complications in other works.
Patchwise training is common \cite{ning2005automatic, Ciresan, Farabet, Pinheiro, N4}, but lacks the efficiency of fully convolutional training.
Our approach does not make use of pre- and post-processing complications, including superpixels \cite{Farabet, Bharath}, proposals \cite{Bharath, Saurabh}, or post-hoc refinement by random fields or local classifiers \cite{Farabet, Bharath}.
Our model transfers recent success in classification \cite{AlexNet, VGGNet, GoogLeNet} to dense prediction by reinterpreting classification nets as fully convolutional and fine-tuning from their learned representations.
In contrast, previous works have applied small convnets without supervised pre-training \cite{Farabet, Pinheiro, ning2005automatic}.

Semantic segmentation faces an inherent tension between semantics and location: global information resolves \emph{what} while local information resolves \emph{where}.
What can be done to navigate this spectrum from location to semantics?
How can local decisions respect global structure?
It is not immediately clear that deep networks for image classification yield representations sufficient for accurate, pixelwise recognition.

In the conference version of this paper \cite{fcn}, we cast pre-trained networks into fully convolutional form, and augment them with a skip architecture that takes advantage of the full feature spectrum.
The skip architecture fuses the feature hierarchy to combine deep, coarse, semantic information and shallow, fine, appearance information (see Section \ref{sec:skip} and Figure \ref{fig:nets}).
In this light, deep feature hierarchies encode location and semantics in a nonlinear local-to-global pyramid.

This journal paper extends our earlier work \cite{fcn} through further tuning, analysis, and more results.
Alternative choices, ablations, and implementation details better cover the space of \NAME s.
Tuning optimization leads to more accurate networks and a means to learn skip architectures all-at-once instead of in stages.
Experiments that mask foreground and background investigate the role of context and shape.
Results on the object and scene labeling of PASCAL-Context reinforce merging object segmentation and scene parsing as unified pixelwise prediction.

In the next section, we review related work on deep classification nets, \NAME s, recent approaches to semantic segmentation using convnets, and extensions to \NAME s.
The following sections explain \NAME\ design, introduce our architecture with in-network upsampling and skip layers, and describe our experimental framework.
Next, we demonstrate improved accuracy on PASCAL VOC 2011-2, NYUDv2, SIFT Flow, and PASCAL-Context.
Finally, we analyze design choices, examine what cues can be learned by an \NAME, and calculate recognition bounds for semantic segmentation.

\begin{figure}
\centering
\includegraphics[width=0.45\textwidth]{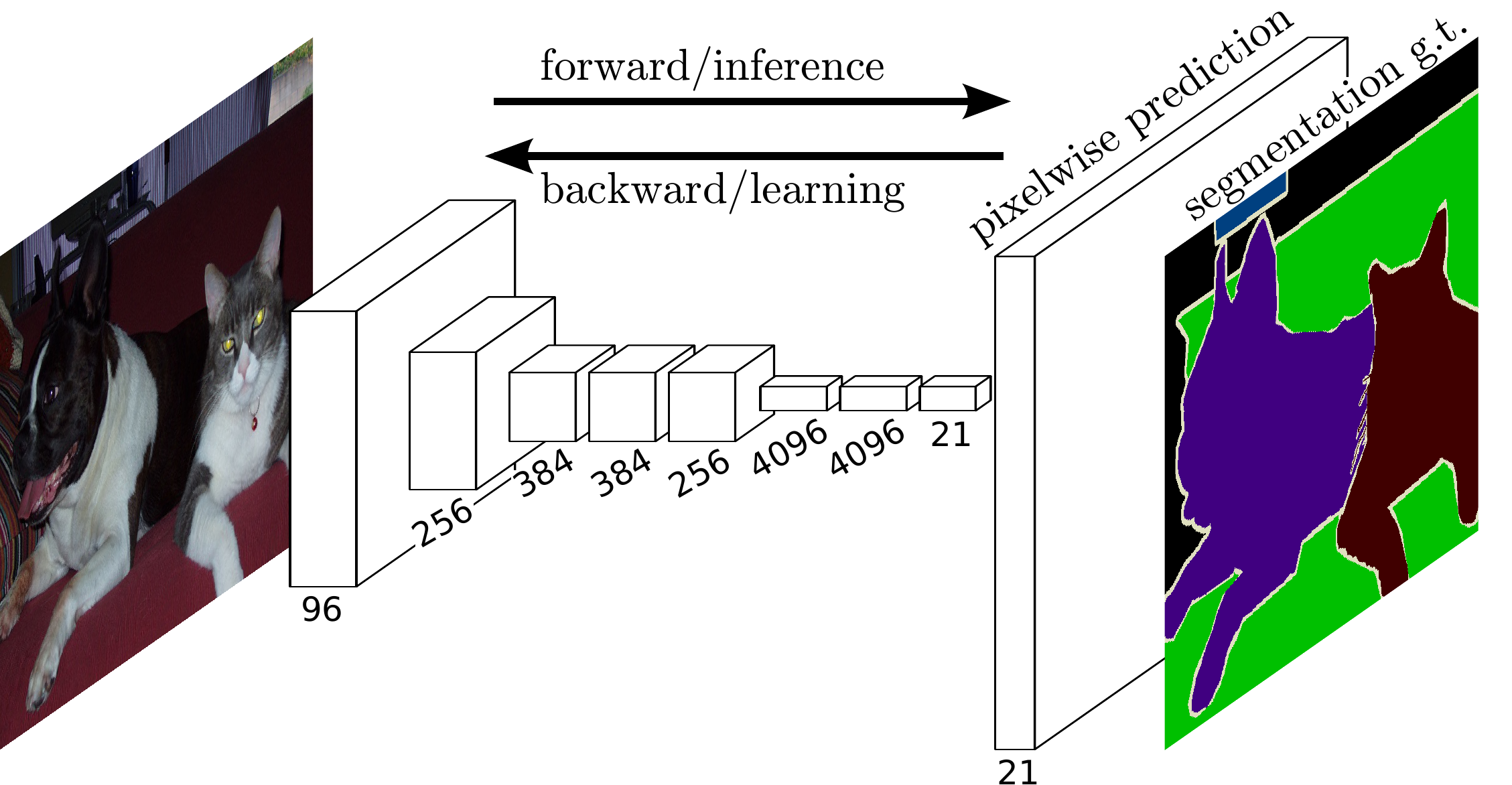}
\caption{
Fully convolutional networks can efficiently learn to make dense predictions for per-pixel tasks like semantic segmentation.
}
\label{fig:model}
\end{figure}

\section{Related Work}

Our approach draws on recent successes of deep nets for image classification \cite{AlexNet, VGGNet, GoogLeNet} and transfer learning \cite{decaf,ZF}.
Transfer was first demonstrated on various visual recognition tasks \cite{decaf,ZF}, then on detection, and on both instance and semantic segmentation in hybrid proposal-classifier models \cite{RCNN, Bharath, Saurabh}.
We now re-architect and fine-tune classification nets to direct, dense prediction of semantic segmentation.
We chart the space of \NAME s and relate prior models both historical and recent.

\minisection{\Name s}
To our knowledge, the idea of extending a convnet to arbitrary-sized inputs first appeared in Matan \etal \cite{Matan}, which extended the classic LeNet \cite{LeNet} to recognize strings of digits.
Because their net was limited to one-dimensional input strings, Matan \etal used Viterbi decoding to obtain their outputs.
Wolf and Platt \cite{wolf1994postal} expand convnet outputs to 2-dimensional maps of detection scores for the four corners of postal address blocks.
Both of these historical works do inference and learning fully convolutionally for detection.
Ning \etal \cite{ning2005automatic} define a convnet for coarse multiclass segmentation of \emph{C. elegans} tissues with fully convolutional inference.

Fully convolutional computation has also been exploited in the present era of many-layered nets.
Sliding window detection by Sermanet \etal \cite{OverFeat}, semantic segmentation by Pinheiro and Collobert \cite{Pinheiro}, and image restoration by Eigen \etal \cite{Eigenrestore} do fully convolutional inference.
Fully convolutional training is rare, but used effectively by Tompson \etal \cite{Tompson} to learn an end-to-end part detector and spatial model for pose estimation, although they do not exposit on or analyze this method.

\minisection{Dense prediction with convnets}
Several recent works have applied convnets to dense prediction problems,
including semantic segmentation by Ning \etal \cite{ning2005automatic}, Farabet \etal \cite{Farabet}, and Pinheiro and Collobert \cite{Pinheiro};
boundary prediction for electron microscopy by Ciresan \etal \cite{Ciresan} and for natural images by a hybrid convnet/nearest neighbor model by Ganin and Lempitsky \cite{N4};
and image restoration and depth estimation by Eigen \etal \cite{Eigenrestore, Eigendepth}.
Common elements of these approaches include
\begin{itemize}[noitemsep,topsep=0pt,parsep=0pt,partopsep=0pt,leftmargin=10pt,itemindent=0pt]
  \item small models restricting capacity and receptive fields;
  \item patchwise training \cite{ning2005automatic, Ciresan, Farabet, Pinheiro, N4};
  \item refinement by superpixel projection, random field regularization, filtering, or local classification \cite{Farabet, Ciresan, N4};
  \item ``interlacing'' to obtain dense output \cite{OverFeat, Pinheiro, N4};
  \item multi-scale pyramid processing \cite{Farabet, Pinheiro, N4};
  \item saturating $\tanh$ nonlinearities \cite{Farabet, Eigenrestore, Pinheiro}; and
  \item ensembles \cite{Ciresan, N4},
\end{itemize}
whereas our method does without this machinery.
However, we do study patchwise training (Section \ref{sec:patches}) and ``shift-and-stitch'' dense output (Section \ref{sec:shifting}) from the perspective of \NAME s.
We also discuss in-network upsampling (Section \ref{sec:upsampling}), of which the fully connected prediction by Eigen \etal \cite{Eigendepth} is a special case.

Unlike these existing methods, we adapt and extend deep classification architectures, using image classification as supervised pre-training, and fine-tune fully convolutionally to learn simply and efficiently from whole image inputs and whole image ground thruths.

Hariharan \etal \cite{Bharath} and Gupta \etal \cite{Saurabh} likewise adapt deep classification nets to semantic segmentation, but do so in hybrid proposal-classifier models.
These approaches fine-tune an R-CNN system \cite{RCNN} by sampling bounding boxes and/or region proposals for detection, semantic segmentation, and instance segmentation.
Neither method is learned end-to-end.
They achieve the previous best segmentation results on PASCAL VOC and NYUDv2 respectively, so we directly compare our standalone, end-to-end \NAME\ to their semantic segmentation results in Section \ref{sec:results}.

\minisection{Combining feature hierarchies}
We fuse features across layers to define a nonlinear local-to-global representation that we tune end-to-end.
The Laplacian pyramid \cite{burt1983laplacian} is a classic multi-scale representation made of fixed smoothing and differencing.
The jet of Koenderink and van Doorn \cite{koenderink1987representation} is a rich, local feature defined by compositions of partial derivatives.
In the context of deep networks, Sermanet \etal \cite{sermanet-cvpr13} fuse intermediate layers but discard resolution in doing so.
In contemporary work Hariharan \etal \cite{NewBharath} and Mostajabi \etal \cite{zoomout} also fuse multiple layers but do not learn end-to-end and rely on fixed bottom-up grouping.

\minisection{\NAME\ extensions} Following the conference version of this paper \cite{fcn}, \NAME s have been extended to new tasks and data.
Tasks include region proposals \cite{ren2015faster}, contour detection \cite{xie2015holistically}, depth regression \cite{liu2015learning}, optical flow \cite{fischer2015flownet}, and weakly-supervised semantic segmentation \cite{Deepak, papandreou2015weakly, dai2015boxsup, hong2015decoupled}.

In addition, new works have improved the \NAME s presented here to further advance the state-of-the-art in semantic segmentation.
The DeepLab models \cite{deeplab} raise output resolution by dilated convolution and dense CRF inference.
The joint CRFasRNN \cite{zheng2015conditional} model is an end-to-end integration of the CRF for further improvement.
ParseNet \cite{liu2015parsenet} normalizes features for fusion and captures context with global pooling.
The ``deconvolutional network'' approach of \cite{noh2015learning} restores resolution by proposals, stacks of learned deconvolution, and unpooling.
U-Net \cite{ronneberger2015u} combines skip layers and learned deconvolution for pixel labeling of microscopy images.
The dilation architecture of \cite{yu2015multi} makes thorough use of dilated convolution for pixel-precise output without a random field or skip layers.

\section{Fully Convolutional Networks}
\label{sec:fc}

Each layer output in a convnet is a three-dimensional array of size $h \times w \times d$, where $h$ and $w$ are spatial dimensions, and $d$ is the feature or channel dimension.
The first layer is the image, with pixel size $h \times w$, and $d$ channels.
Locations in higher layers correspond to the locations in the image they are path-connected to, which are called their \emph{receptive fields}.

Convnets are inherently translation invariant.
Their basic components (convolution, pooling, and activation functions) operate on local input regions, and depend only on \emph{relative} spatial coordinates.
Writing $\mathbf x_{ij}$ for the data vector at location $(i, j)$ in a particular layer, and $\mathbf y_{ij}$ for the following layer, these functions compute outputs $\mathbf y_{ij}$ by
\[
\mathbf y_{ij} = f_{ks}\left(
\{\mathbf x_{si + \delta i, sj + \delta j}\}_{0 \leq \delta i,
\delta j < k}\right)
\]
where $k$ is called the kernel size, $s$ is the stride or subsampling factor, and $f_{ks}$ determines the layer type: a matrix multiplication for convolution or average pooling, a spatial max for max pooling, or an elementwise nonlinearity for an activation function, and so on for other types of layers.

This functional form is maintained under composition, with kernel size and stride obeying the transformation rule
\[
f_{ks} \circ g_{k's'} = (f \circ g)_{k'+(k-1)s', ss'}.
\]
While a general net computes a general nonlinear function, a net with only layers of this form computes a nonlinear \emph{filter}, which we call a \emph{deep filter} or \emph{fully convolutional network}.
An \NAME\ naturally operates on an input of any size, and produces an output of corresponding (possibly resampled) spatial dimensions.

A real-valued loss function composed with an \NAME\ defines a task.
If the loss function is a sum over the spatial dimensions of the final layer, $\ell(\mathbf x; \theta) = \sum_{ij} \ell'(\mathbf x_{ij}; \theta)$, its parameter gradient will be a sum over the parameter gradients of each of its spatial components.
Thus stochastic gradient descent on $\ell$ computed on whole images will be the same as stochastic gradient descent on $\ell'$, taking all of the final layer receptive fields as a minibatch.

When these receptive fields overlap significantly, both feedforward computation \emph{and} backpropagation are much more efficient when computed layer-by-layer over an entire image instead of independently patch-by-patch.

We next explain how to convert classification nets into fully convolutional nets that produce coarse output maps.
For pixelwise prediction, we need to connect these coarse outputs back to the pixels.
Section \ref{sec:shifting} describes a trick used for this purpose (e.g., by ``fast scanning'' \cite{giusti2013fast}).
We explain this trick in terms of network modification.
As an efficient, effective alternative, we upsample in Section \ref{sec:upsampling}, reusing our implementation of convolution.
In Section \ref{sec:patches} we consider training by patchwise sampling, and give evidence in Section \ref{sec:training} that our whole image training is faster and equally effective.

\subsection{Adapting classifiers for dense prediction}
\label{sec:adapting}

\begin{figure}
\includegraphics[width=0.45\textwidth]{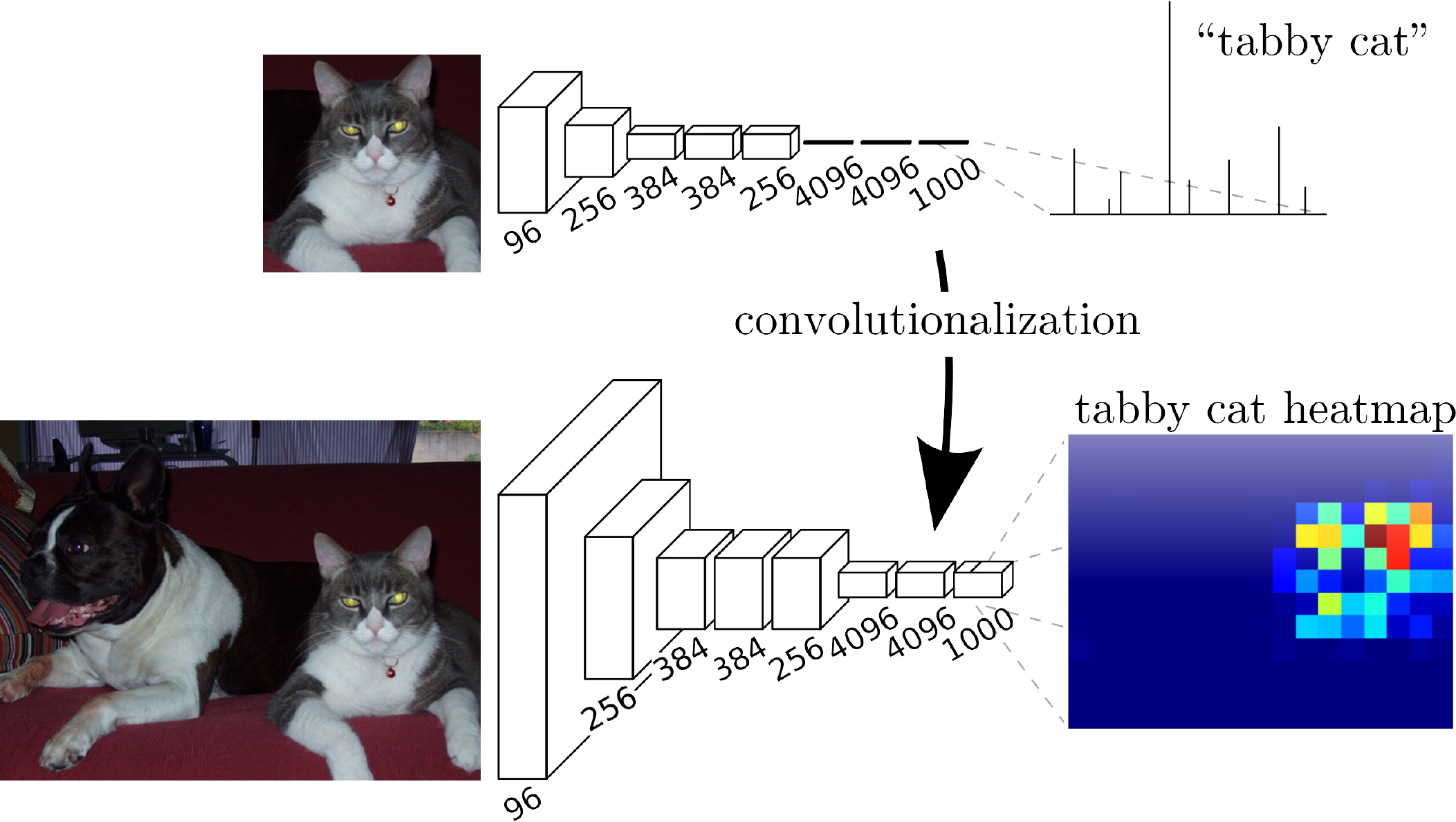}
\caption{
Transforming fully connected layers into convolution layers enables a classification net to output a spatial map.
Adding differentiable interpolation layers and a spatial loss (as in Figure \ref{fig:model}) produces an efficient machine for end-to-end pixelwise learning.
}
\label{fig:alex-evo}
\end{figure}

Typical recognition nets, including LeNet \cite{LeNet}, AlexNet \cite{AlexNet}, and its deeper successors \cite{VGGNet, GoogLeNet}, ostensibly take fixed-sized inputs and produce non-spatial outputs.
The fully connected layers of these nets have fixed dimensions and throw away spatial coordinates.
However, fully connected layers can also be viewed as convolutions with kernels that cover their entire input regions.
Doing so casts these nets into \name s that take input of any size and make spatial output maps.
This transformation is illustrated in Figure \ref{fig:alex-evo}.

Furthermore, while the resulting maps are equivalent to the evaluation of the original net on particular input patches, the computation is highly amortized over the overlapping regions of those patches.
For example, while AlexNet takes $1.2$ ms (on a typical GPU) to infer the classification scores of a $227 \times 227$ image, the fully convolutional net takes $22$ ms to produce a $10 \times 10$ grid of outputs from a $500 \times 500$ image, which is more than $5$ times faster than the na\"ive approach\footnote{Assuming efficient batching of single image inputs.
The classification scores for a single image by itself take 5.4 ms to produce, which is nearly $25$ times slower than the fully convolutional version.}.

The spatial output maps of these convolutionalized models make them a natural choice for dense problems like semantic segmentation.
With ground truth available at every output cell, both the forward and backward passes are straightforward, and both take advantage of the inherent computational efficiency (and aggressive optimization) of convolution.
The corresponding backward times for the AlexNet example are $2.4$ ms for a single image and $37$ ms for a fully convolutional $10 \times 10$ output map, resulting in a speedup similar to that of the forward pass.

While our reinterpretation of classification nets as fully convolutional yields output maps for inputs of any size, the output dimensions are typically reduced by subsampling.
The classification nets subsample to keep filters small and computational requirements reasonable.
This coarsens the output of a fully convolutional version of these nets, reducing it from the size of the input by a factor equal to the pixel stride of the receptive fields of the output units.

\subsection{Shift-and-stitch is filter dilation}
\label{sec:shifting}

Dense predictions can be obtained from coarse outputs by stitching together outputs from shifted versions of the input.
If the output is downsampled by a factor of $f$, shift the input $x$ pixels to the right and $y$ pixels down, once for every $(x, y)$ such that
$0 \leq x, y < f$.
Process each of these $f^2$ inputs, and interlace the outputs so that the predictions correspond to the pixels at the \emph{centers} of their receptive fields.

Although this transformation na\"ively increases the cost by a factor of $f^2$, there is a well-known trick for efficiently producing identical results \cite{giusti2013fast, OverFeat}.
(This trick is also used in the algorithme \`a trous \cite{holschneider87, mallat1999wavelet} for wavelet transforms and related to the Noble identities \cite{signals} from signal processing.)

Consider a layer (convolution or pooling) with input stride $s$, and a subsequent convolution layer with filter weights $f_{ij}$ (eliding the irrelevant feature dimensions).
Setting the earlier layer's input stride to one upsamples its output by a factor of $s$.
However, convolving the original filter with the upsampled output does not produce the same result as shift-and-stitch, because the original filter only sees a reduced portion of its (now upsampled) input.
To produce the same result, dilate (or ``rarefy'') the filter by forming
\[
f'_{ij} = \left\{
\begin{tabular}{ll}
$f_{i/s, j/s}$ & if $s$ divides both $i$ and $j$; \\
$0$ & otherwise,
\end{tabular}
\right.
\]
(with $i$ and $j$ zero-based).
Reproducing the full net output of shift-and-stitch involves repeating this filter enlargement layer-by-layer until all subsampling is removed. (In practice, this can be done efficiently by processing subsampled versions of the upsampled input.)

Simply decreasing subsampling within a net is a tradeoff: the filters see finer information, but have smaller receptive fields and take longer to compute.
This dilation trick is another kind of tradeoff: the output is denser without decreasing the receptive field sizes of the filters, but the filters are prohibited from accessing information at a finer scale than their original design.

Although we have done preliminary experiments with dilation, we do not use it in our model.
We find learning through upsampling, as described in the next section, to be effective and efficient, especially when combined with the skip layer fusion described later on.
For further detail regarding dilation, refer to the dilated FCN of \cite{yu2015multi}.

\subsection{Upsampling is (fractionally strided) convolution}
\label{sec:upsampling}

Another way to connect coarse outputs to dense pixels is interpolation.
For instance, simple bilinear interpolation computes each output $y_{ij}$ from the nearest four inputs by a linear map that depends only on the relative positions of the input and output cells:
\[
y_{ij} = \sum_{\alpha, \beta=0}^1 |1 - \alpha - \{i/f\}| ~ |1 - \beta - \{i/j\}|
~x_{\lfloor i/f \rfloor + \alpha, \lfloor j/f \rfloor + \beta},
\]
where $f$ is the upsampling factor, and $\{\cdot\}$ denotes the fractional part.

In a sense, upsampling with factor $f$ is convolution with a \emph{fractional} input stride of $1/f$.
So long as $f$ is integral, it's natural to implement upsampling through ``backward convolution'' by reversing the forward and backward passes of more typical input-strided convolution.
Thus upsampling is performed in-network for end-to-end learning by backpropagation from the pixelwise loss.

Per their use in deconvolution networks (esp.\ \cite{ZF}), these (convolution) layers are sometimes referred to as \emph{deconvolution} layers.
Note that the convolution filter in such a layer need not be fixed (e.g., to bilinear upsampling), but can be learned.
A stack of deconvolution layers and activation functions can even learn a nonlinear upsampling.

In our experiments, we find that in-network upsampling is fast and effective for learning dense prediction.

\subsection{Patchwise training is loss sampling}
\label{sec:patches}

In stochastic optimization, gradient computation is driven by the training distribution.
Both patchwise training and fully convolutional training can be made to produce any distribution of the inputs, although their relative computational efficiency depends on overlap and minibatch size.
Whole image fully convolutional training is identical to patchwise training where each batch consists of all the receptive fields of the output units for an image (or collection of images).
While this is more efficient than uniform sampling of patches, it reduces the number of possible batches.
However, random sampling of patches within an image may be easily recovered.
Restricting the loss to a randomly sampled subset of its spatial terms (or, equivalently applying a DropConnect mask \cite{dropconnect} between the output and the loss) excludes patches from the gradient.

If the kept patches still have significant overlap, fully convolutional computation will still speed up training.
If gradients are accumulated over multiple backward passes, batches can include patches from several images.
If inputs are shifted by values up to the output stride, random selection of all possible patches is possible even though the output units lie on a fixed, strided grid.

Sampling in patchwise training can correct class imbalance \cite{ning2005automatic, Farabet, Ciresan} and mitigate the spatial correlation of dense patches \cite{Pinheiro, Bharath}.
In fully convolutional training, class balance can also be achieved by weighting the loss, and loss sampling can be used to address spatial correlation.

We explore training with sampling in Section \ref{sec:training}, and do not find that it yields faster or better convergence for dense prediction. Whole image training is effective and efficient.

\section{Segmentation Architecture}
\label{sec:arch}

We cast ILSVRC classifiers into \NAME s and augment them for dense prediction with in-network upsampling and a pixelwise loss.
We train for segmentation by fine-tuning.
Next, we add skips between layers to fuse coarse, semantic and local, appearance information.
This skip architecture is learned end-to-end to refine the semantics and spatial precision of the output.

For this investigation, we train and validate on the PASCAL VOC 2011 segmentation challenge \cite{PASCAL}.
We train with a per-pixel softmax loss and validate with the standard metric of mean pixel intersection over union, with the mean taken over all classes, including background.
The training ignores pixels that are masked out (as ambiguous or difficult) in the ground truth.

\subsection{From classifier to dense \NAME}
\label{sec:base}

\begin{table}[t]
\centering
\caption{
We adapt and extend three classification convnets.
We compare performance by mean intersection over union on the validation set of PASCAL VOC 2011 and by inference time (averaged over 20 trials for a $500 \times 500$ input on an NVIDIA Titan X).
We detail the architecture of the adapted nets with regard to dense prediction: number of parameter layers, receptive field size of output units, and the coarsest stride within the net.
(These numbers give the best performance obtained at a fixed learning rate, not best performance possible.)
}
\setlength{\tabcolsep}{2pt}
\begin{tabular}{llll}
\toprule
& FCN-AlexNet & FCN-VGG16 & FCN-GoogLeNet\footnotemark[3] \\
\midrule
mean IU & 39.8 & \textbf{56.0} & 42.5 \\
forward time & 16 ms & 100 ms & 20 ms \\
conv.\ layers & 8 & 16 & 22 \\
parameters & 57M & 134M & 6M \\
rf size & 355 & 404 & 907 \\
max stride & 32 & 32 & 32 \\
\bottomrule
\end{tabular}
\label{tab:base}
\end{table}

We begin by convolutionalizing proven classification architectures as in Section \ref{sec:fc}.
We consider the AlexNet\footnote{Using the publicly available \texttt{CaffeNet} reference model.} architecture \cite{AlexNet} that won ILSVRC12, as well as the VGG nets \cite{VGGNet} and the GoogLeNet%
\footnote{%
We use our own reimplementation of GoogLeNet.
Ours is trained with less extensive data augmentation, and gets 68.5\% top-1 and 88.4\% top-5 ILSVRC accuracy.} \cite{GoogLeNet} which did exceptionally well in ILSVRC14.
We pick the VGG 16-layer net\footnote{Using the publicly available version from the Caffe model zoo.}, which we found to be equivalent to the 19-layer net on this task.
For GoogLeNet, we use only the final loss layer, and improve performance by discarding the final average pooling layer.
We decapitate each net by discarding the final classifier layer, and convert all fully connected layers to convolutions.
We append a $1 \times 1$ convolution with channel dimension $21$ to predict scores for each of the PASCAL classes (including background) at each of the coarse output locations, followed by a (backward) convolution layer to bilinearly upsample the coarse outputs to pixelwise outputs as described in Section \ref{sec:upsampling}.
Table \ref{tab:base} compares the preliminary validation results along with the basic characteristics of each net.
We report the best results achieved after convergence at a fixed learning rate (at least 175 epochs).

Our training for this comparison follows the practices for classification networks.
We train by SGD with momentum.
Gradients are accumulated over 20 images.
We set fixed learning rates of $10^{-3}$, $10^{-4}$, and $5^{-5}$ for FCN-AlexNet, FCN-VGG16, and FCN-GoogLeNet, respectively, chosen by line search.
We use momentum $0.9$, weight decay of $5^{-4}$ or $2^{-4}$, and doubled learning rate for biases.
We zero-initialize the class scoring layer, as random initialization yielded neither better performance nor faster convergence.
Dropout is included where used in the original classifier nets (however, training without it made little to no difference).

Fine-tuning from classification to segmentation gives reasonable predictions from each net.
Even the worst model achieved $\sim 75\%$ of the previous best performance.
FCN-VGG16 already appears to be better than previous methods at 56.0 mean IU on val, compared to 52.6 on test \cite{Bharath}.
Although VGG and GoogLeNet are similarly accurate as classifiers, our FCN-GoogLeNet did not match FCN-VGG16.
We select FCN-VGG16 as our base network.

\subsection{Image-to-image learning}
\label{sec:opt}

\begin{table}[t]
\centering
\caption{
Comparison of image-to-image optimization by gradient accumulation, online learning, and ``heavy'' learning with high momentum.
All methods are trained on a fixed sequence of 100,000 images (sampled from a dataset of 8,498) to control for stochasticity and equalize the number of gradient computations.
The loss is not normalized so that every pixel has the same weight no matter the batch and image dimensions.
Scores are the best achieved during training on a subset\protect\footnotemark[5] of PASCAL VOC 2011 segval.
Learning is end-to-end with FCN-VGG16.
}
\begin{tabular}{lcccccc}
\toprule
& \metr{batch\\size} & mom. & \pixacc & \classacc & \meanIU & \fwIU \\
\midrule
FCN-accum  & 20 & 0.9 & 86.0 & 66.5 & 51.9 & 76.5 \\
FCN-online & 1 & 0.9 & 89.3 & 76.2 & 60.7 & 81.8 \\
FCN-heavy & 1 & 0.99 & \textbf{90.5} & \textbf{76.5} & \textbf{63.6} & \textbf{83.5} \\
\bottomrule
\end{tabular}
\label{tab:online}
\end{table}

The image-to-image learning setting includes high effective batch size and correlated inputs.
This optimization requires some attention to properly tune \NAME s.

We begin with the loss.
We do not normalize the loss, so that every pixel has the same weight regardless of the batch and image dimensions.
Thus we use a small learning rate since the loss is summed spatially over all pixels.

We consider two regimes for batch size.
In the first, gradients are accumulated over 20 images.
Accumulation reduces the memory required and respects the different dimensions of each input by reshaping the network.
We picked this batch size empirically to result in reasonable convergence.
Learning in this way is similar to standard classification training: each minibatch contains several images and has a varied distribution of class labels.
The nets compared in Table \ref{tab:base} are optimized in this fashion.

However, batching is not the only way to do image-wise learning.
In the second regime, batch size \emph{one} is used for online learning.
Properly tuned, online learning achieves higher accuracy and faster convergence in both number of iterations and wall clock time.
Additionally, we try a higher momentum of $0.99$, which increases the weight on recent gradients in a similar way to batching.
See Table \ref{tab:online} for the comparison of accumulation, online, and high momentum or ``heavy'' learning (discussed further in Section \ref{ana:opt}).

\subsection{Combining \emph{what} and \emph{where}}
\label{sec:skip}

\begin{figure*}[tp]
\includegraphics[width=\textwidth]{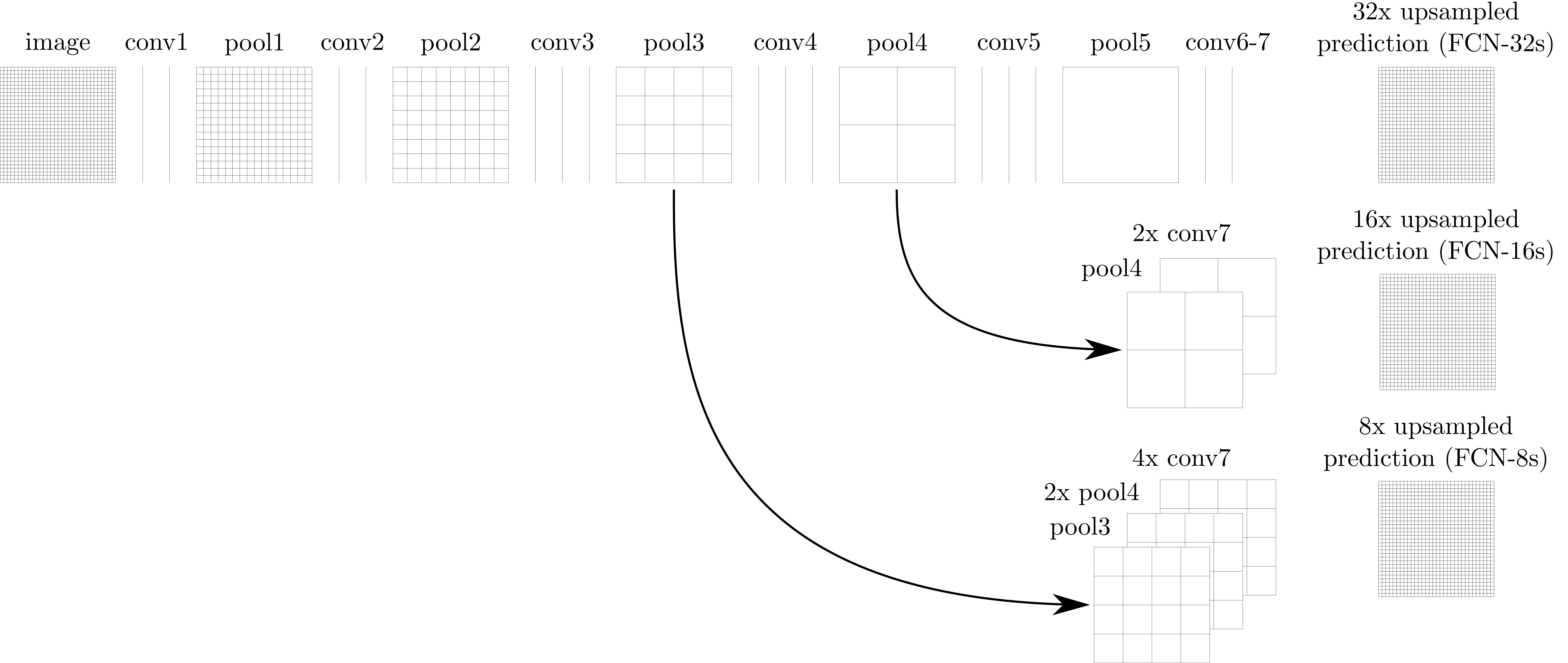}
\caption{
Our DAG nets learn to combine coarse, high layer information with fine, low layer information.
Pooling and prediction layers are shown as grids that reveal relative spatial coarseness, while intermediate layers are shown as vertical lines.
First row (FCN-32s): Our single-stream net, described in Section \ref{sec:base}, upsamples stride 32 predictions back to pixels in a single step.
Second row (FCN-16s): Combining predictions from both the final layer and the \texttt{pool4} layer, at stride 16, lets our net predict finer details, while retaining high-level semantic information.
Third row (FCN-8s): Additional predictions from \texttt{pool3}, at stride 8, provide further precision.
}
\label{fig:nets}
\end{figure*}

We define a new fully convolutional net for segmentation that combines layers of the feature hierarchy and refines the spatial precision of the output.
See Figure \ref{fig:nets}.

\begin{figure}
\centering
\setlength{\tabcolsep}{1pt}
\hspace*{-4pt}
\begin{tabular}{cccc}
FCN-32s & FCN-16s & FCN-8s & Ground truth \\
\includegraphics[width=0.12\textwidth]{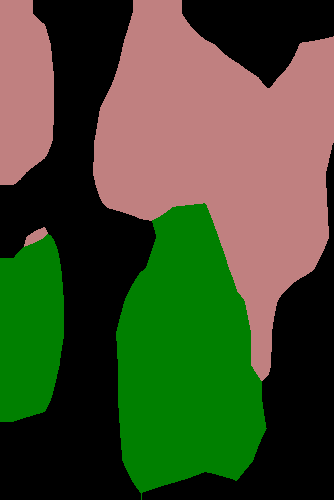} &
\includegraphics[width=0.12\textwidth]{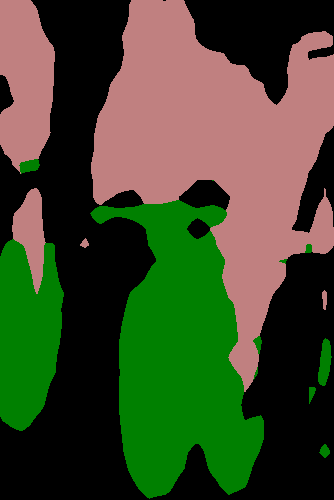} &
\includegraphics[width=0.12\textwidth]{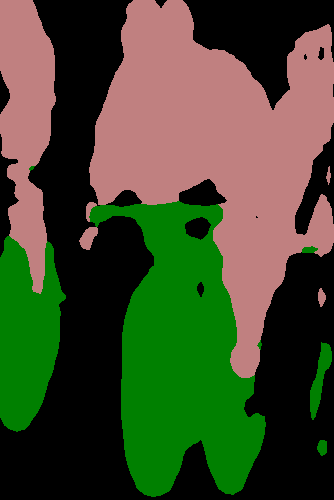} &
\includegraphics[width=0.12\textwidth]{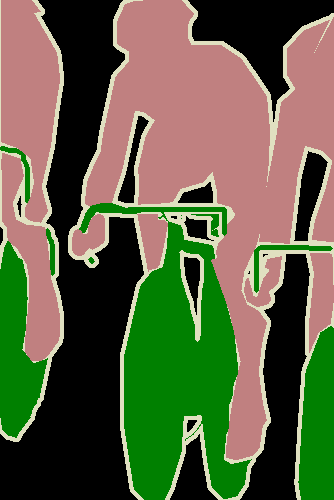}
\end{tabular}
\caption{
Refining fully convolutional networks by fusing information from layers with different strides improves spatial detail.
The first three images show the output from our 32, 16, and 8 pixel stride nets (see Figure \ref{fig:nets}).
}
\label{fig:evo}
\end{figure}

While fully convolutionalized classifiers fine-tuned to semantic segmentation both recognize and localize, as shown in Section \ref{sec:base}, these networks can be improved to make direct use of shallower, more local features.
Even though these base networks score highly on the standard metrics, their output is dissatisfyingly coarse (see Figure \ref{fig:evo}).
The stride of the network prediction limits the scale of detail in the upsampled output.

We address this by adding skips \cite{bishop2006pattern} that fuse layer outputs, in particular to include shallower layers with finer strides in prediction.
This turns a line topology into a DAG: edges skip ahead from shallower to deeper layers.
It is natural to make more local predictions from shallower layers since their receptive fields are smaller and see fewer pixels.
Once augmented with skips, the network makes and fuses predictions from several streams that are learned jointly and end-to-end.

Combining fine layers and coarse layers lets the model make local predictions that respect global structure.
This crossing of layers and resolutions is a learned, nonlinear counterpart to the multi-scale representation of the Laplacian pyramid \cite{burt1983laplacian}.
By analogy to the jet of Koenderick and van Doorn \cite{koenderink1987representation}, we call our feature hierarchy the \emph{deep jet}.

Layer fusion is essentially an elementwise operation.
However, the correspondence of elements across layers is complicated by resampling and padding.
Thus, in general, layers to be fused must be aligned by scaling and cropping.
We bring two layers into scale agreement by upsampling the lower-resolution layer, doing so in-network as explained in Section \ref{sec:upsampling}.
Cropping removes any portion of the upsampled layer which extends beyond the other layer due to padding.
This results in layers of equal dimensions in exact alignment.
The offset of the cropped region depends on the resampling and padding parameters of all intermediate layers.
Determining the crop that results in exact correspondence can be intricate, but it follows automatically from the network definition (and we include code for it in Caffe).

Having spatially aligned the layers, we next pick a fusion operation.
We fuse features by concatenation, and immediately follow with classification by a ``score layer'' consisting of a $1 \times 1$ convolution.
Rather than storing concatenated features in memory, we commute the concatenation and subsequent classification (as both are linear).
Thus, our skips are implemented by first scoring each layer to be fused by $1 \times 1$ convolution, carrying out any necessary interpolation and alignment, and then \emph{summing} the scores.
We also considered max fusion, but found learning to be difficult due to gradient switching.
The score layer parameters are zero-initialized when a skip is added, so that they do not interfere with existing predictions of other streams.
Once all layers have been fused, the final prediction is then upsampled back to image resolution.

\minisection{Skip Architectures for Segmentation}
We define a skip architecture to extend FCN-VGG16 to a three-stream net with eight pixel stride shown in Figure \ref{fig:nets}.
Adding a skip from \texttt{pool4} halves the stride by scoring from this stride sixteen layer.
The $2 \times$ interpolation layer of the skip is initialized to bilinear interpolation, but is not fixed so that it can be learned as described in Section \ref{sec:upsampling}.
We call this two-stream net FCN-16s, and likewise define FCN-8s by adding a further skip from \texttt{pool3} to make stride eight predictions.
(Note that predicting at stride eight does not significantly limit the maximum achievable mean IU; see Section \ref{sec:ub}.)

We experiment with both \emph{staged training} and \emph{all-at-once training}.
In the staged version, we learn the single-stream FCN-32s, then upgrade to the two-stream FCN-16s and continue learning, and finally upgrade to the three-stream FCN-8s and finish learning.
At each stage the net is learned end-to-end, initialized with the parameters of the earlier net.
The learning rate is dropped $100\times$ from FCN-32s to FCN-16s and $100\times$ more from FCN-16s to FCN-8s, which we found to be necessary for continued improvements.

Learning all-at-once rather than in stages gives nearly equivalent results, while training is faster and less tedious.
However, disparate feature scales make na\"ive training prone to divergence.
To remedy this we scale each stream by a fixed constant, for a similar in-network effect to the staged learning rate adjustments.
These constants are picked to approximately equalize average feature norms across streams.
(Other normalization schemes should have similar effect.)

With FCN-16s validation score improves to 65.0 mean IU, and FCN-8s brings a minor improvement to 65.5.
At this point our fusion improvements have met diminishing returns,
so we do not continue fusing even shallower layers.

To identify the contribution of the skips we compare scoring from the intermediate layers in isolation, which results in poor performance, or dropping the learning rate without adding skips, which gives negligible improvement in score without refining the visual quality of output.
All skip comparisons are reported in Table \ref{tab:fcnarch}.
Figure \ref{fig:evo} shows the progressively finer structure of the output.

\begin{table}[h!]
\centering
\caption{
Comparison of \NAME s on a subset\protect\footnotemark[5] of PASCAL VOC 2011 segval.
Learning is end-to-end with batch size one and high momentum, with the exception of the fixed variant that fixes all features.
Note that FCN-32s is FCN-VGG16, renamed to highlight stride, and the FCN-poolX are truncated nets with the same strides as FCN-32/16/8s.
}
\begin{tabular}{lcccc}
\toprule
& \pixacc & \classacc & \meanIU & \fwIU \\
\midrule
FCN-32s & 90.5 & 76.5 & 63.6 & 83.5 \\
FCN-16s & 91.0 & 78.1 & 65.0 & 84.3 \\
FCN-8s at-once & 91.1 & \textbf{78.5} & 65.4 & 84.4 \\
FCN-8s staged  & \textbf{91.2} & 77.6 & \textbf{65.5} & \textbf{84.5} \\
\midrule
FCN-32s fixed    & 82.9 & 64.6 & 46.6 & 72.3 \\
\midrule
FCN-pool5     & 87.4 & 60.5 & 50.0 & 78.5 \\
FCN-pool4     & 78.7 & 31.7 & 22.4 & 67.0 \\
FCN-pool3     & 70.9 & 13.7 &  9.2 & 57.6 \\
\bottomrule
\end{tabular}
\label{tab:fcnarch}
\end{table}

\subsection{Experimental framework}
\label{sec:training}

\minisection{Fine-tuning}
We fine-tune all layers by backpropagation through the whole net.
Fine-tuning the output classifier alone yields only 73\% of the full fine-tuning performance as compared in Table \ref{tab:fcnarch}.
Fine-tuning in stages takes 36 hours on a single GPU.
Learning FCN-8s all-at-once takes half the time to reach comparable accuracy.
Training from scratch gives substantially lower accuracy.

\minisection{More training data}
The PASCAL VOC 2011 segmentation training set labels 1,112 images.
Hariharan \etal \cite{BharathData} collected labels for a larger set of 8,498 PASCAL training images, which was used to train the previous best system, SDS \cite{Bharath}.
This training data improves the FCN-32s validation score%
\footnote{There are training images from \cite{BharathData} included in the PASCAL VOC 2011 val set, so we validate on the non-intersecting set of 736 images.
}
from 57.7 to 63.6 mean IU and improves the FCN-AlexNet score from 39.8 to 48.0 mean IU.

\minisection{Loss}
The per-pixel, unnormalized softmax loss is a natural choice for segmenting images of any size into disjoint classes, so we train our nets with it.
The softmax operation induces competition between classes and promotes the most confident prediction, but it is not clear that this is necessary or helpful.
For comparison, we train with the sigmoid cross-entropy loss and find that it gives similar results, even though it normalizes each class prediction independently.

\begin{figure}
\centering
\includegraphics[width=0.23\textwidth]{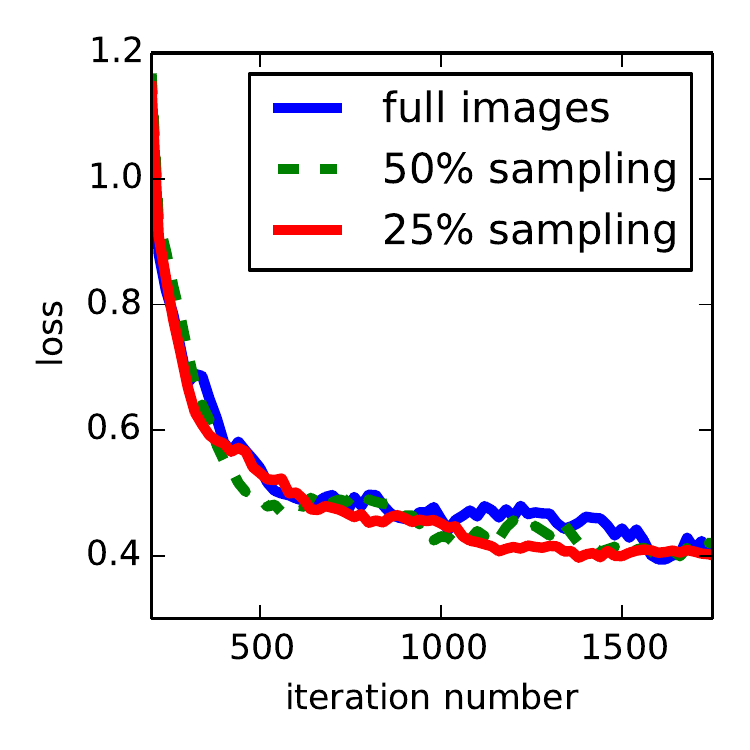}
\includegraphics[width=0.23\textwidth]{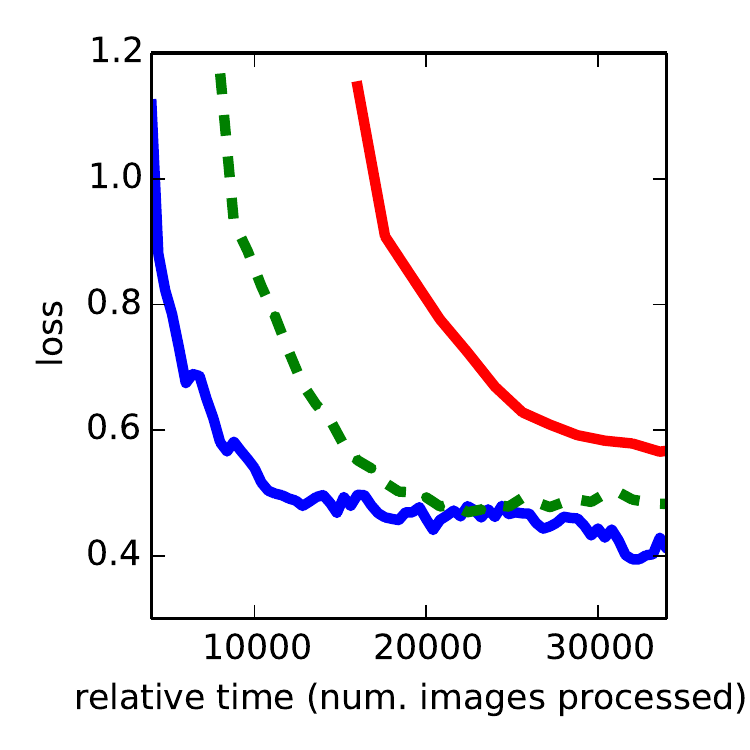}
\caption{
Training on whole images is just as effective as sampling patches, but results in faster (wall clock time) convergence by making more efficient use of data.
Left shows the effect of sampling on convergence rate for a fixed expected batch size, while right plots the same by relative wall clock time.
}
\label{fig:sampling}
\end{figure}

\minisection{Patch sampling}
As explained in Section \ref{sec:patches}, our whole image training effectively batches each image into a regular grid of large, overlapping patches.
By contrast, prior work randomly samples patches over a full dataset \cite{ning2005automatic, Ciresan, Farabet, Pinheiro, N4}, potentially resulting in higher variance batches that may accelerate convergence \cite{backprop}.
We study this tradeoff by spatially sampling the loss in the manner described earlier, making an independent choice to ignore each final layer cell with some probability $1 - p$.
To avoid changing the effective batch size, we simultaneously increase the number of images per batch by a factor $1/p$.
Note that due to the efficiency of convolution, this form of rejection sampling is still faster than patchwise training for large enough values of $p$ (e.g., at least for $p > 0.2$ according to the numbers in Section \ref{sec:adapting}).
Figure \ref{fig:sampling} shows the effect of this form of sampling on convergence.
We find that sampling does not have a significant effect on convergence rate compared to whole image training, but takes significantly more time due to the larger number of images that need to be considered per batch.
We therefore choose unsampled, whole image training in our other experiments.

\minisection{Class balancing}
Fully convolutional training can balance classes by weighting or sampling the loss.
Although our labels are mildly unbalanced (about $3/4$ are background), we find class balancing unnecessary.

\minisection{Dense Prediction}
The scores are upsampled to the input dimensions by backward convolution layers within the net.
Final layer backward convolution weights are fixed to bilinear interpolation, while intermediate upsampling layers are initialized to bilinear interpolation, and then learned.
This simple, end-to-end method is accurate and fast.

\minisection{Augmentation}
We tried augmenting the training data by randomly mirroring and ``jittering'' the images by translating them up to 32 pixels (the coarsest scale of prediction) in each direction.
This yielded no noticeable improvement.

\minisection{Implementation}
All models are trained and tested with Caffe \cite{caffe} on a single NVIDIA Titan X.
Our models and code are publicly available at \url{http://fcn.berkeleyvision.org}.

\section{Results}
\label{sec:results}

We test our FCN on semantic segmentation and scene parsing, exploring PASCAL VOC, NYUDv2, SIFT Flow, and PASCAL-Context.
Although these tasks have historically distinguished between objects and regions, we treat both uniformly as pixel prediction.
We evaluate our FCN skip architecture on each of these datasets, and then extend it to multi-modal input for NYUDv2 and multi-task prediction for the semantic and geometric labels of SIFT Flow.
All experiments follow the same network architecture and optimization settings decided on in Section \ref{sec:arch}.

\minisection{Metrics}
We report metrics from common semantic segmentation and scene parsing evaluations that are variations on pixel accuracy and region intersection over union (IU):
\begin{itemize}[itemsep=0.2em,topsep=0pt,parsep=0pt,partopsep=0pt,leftmargin=10pt]
  \item pixel accuracy: \scalebox{0.8}{$\sum_i n_{ii} / \sum_i t_i$} 
  \item mean accuraccy: \scalebox{0.8}{$(1/n_{\text{cl}}) \sum_i n_{ii}/t_i$} 
  \item mean IU: \scalebox{0.8}{$(1/n_{\text{cl}}) \sum_i n_{ii} /\left(t_i + \sum_j n_{ji} - n_{ii}\right)$} 
  \item frequency weighted IU: \scalebox{0.8}{$\left(\sum_k t_k\right)^{-1} \sum_i t_i n_{ii} /\left(t_i + \sum_j n_{ji} - n_{ii}\right)$} 
\end{itemize}
where $n_{ij}$ is the number of pixels of class $i$ predicted to belong to class $j$, there are $n_{\text{cl}}$ different classes, and $t_i = \sum_j n_{ij}$ is the total number of pixels of class $i$.

\minisection{PASCAL VOC}
Table \ref{tab:pascaltest} gives the performance of our FCN-8s on the test sets of PASCAL VOC 2011 and 2012, and compares it to the previous best, SDS \cite{Bharath}, and the well-known R-CNN \cite{RCNN}.
We achieve the best results on mean IU by 30\% relative.
Inference time is reduced $114\times$ (convnet only, ignoring proposals and refinement) or $286\times$ (overall).

\wordsection{NYUDv2} \cite{NYUDv2} is an RGB-D dataset collected using the Microsoft Kinect.
It has 1,449 RGB-D images, with pixelwise labels that have been coalesced into a 40 class semantic segmentation task by Gupta \etal \cite{Saurabh1}.
We report results on the standard split of 795 training images and 654 testing images.
Table \ref{tab:nyud} gives the performance of several net variations.
First we train our unmodified coarse model (FCN-32s) on RGB images.
To add depth information, we train on a model upgraded to take four-channel RGB-D input (early fusion).
This provides little benefit, perhaps due to similar number of parameters or the difficulty of propagating meaningful gradients all the way through the net.
Following the success of Gupta \etal \cite{Saurabh}, we try the three-dimensional HHA encoding of depth and train a net on just this information.
To effectively combine color and depth, we define a ``late fusion'' of RGB and HHA that averages the final layer scores from both nets and learn the resulting two-stream net end-to-end.
This late fusion RGB-HHA net is the most accurate.

\wordsection{SIFT Flow} is a dataset of 2,688 images with pixel labels for 33 semantic classes (``bridge'', ``mountain'', ``sun''), as well as three geometric classes (``horizontal'', ``vertical'', and ``sky'').
An FCN can naturally learn a joint representation that simultaneously predicts both types of labels.
We learn a two-headed version of FCN-32/16/8s with semantic and geometric prediction layers and losses.
This net performs as well on both tasks as two independently trained nets, while learning and inference are essentially as fast as each independent net by itself.
The results in Table \ref{tab:siftflow}, computed on the standard split into 2,488 training and 200 test images,%
\footnote{Three of the SIFT Flow classes are not present in the test set.
We made predictions across all 33 classes, but only included classes actually present in the test set in our evaluation.
}
show better performance on both tasks.

\begin{table}[h!]
\centering
\caption{
Our FCN gives a 30\% relative improvement on the previous best PASCAL VOC 11/12 test results with faster inference and learning.
}
\begin{tabular}{lccc}
\toprule
& mean IU & mean IU & inference \\
& VOC2011 test & VOC2012 test & time \\
\midrule
R-CNN \cite{RCNN} & 47.9 & - & - \\
SDS \cite{Bharath} & 52.6 & 51.6 & $\sim$ 50 s \\
FCN-8s & \textbf{67.5} & \textbf{67.2} & \bf{$\sim$ 100 ms} \\
\bottomrule
\end{tabular}
\label{tab:pascaltest}
\end{table}

\begin{table}[h!]
\centering
\caption{
Results on NYUDv2.
\textit{RGB-D} is early-fusion of the RGB and depth channels at the input.
\textit{HHA} is the depth embedding of \cite{Saurabh} as horizontal disparity, height above ground, and the angle of the local surface normal with the inferred gravity direction.
\textit{RGB-HHA} is the jointly trained late fusion model that sums RGB and HHA predictions.
}
\begin{tabular}{lcccc}
\toprule
  & \pixacc & \classacc & \meanIU & \fwIU \\
\midrule
Gupta \etal \cite{Saurabh} & 60.3 & - & 28.6 & 47.0 \\
FCN-32s RGB     & 61.8 & 44.7 & 31.6 & 46.0 \\
FCN-32s RGB-D   & 62.1 & 44.8 & 31.7 & 46.3 \\
FCN-32s HHA     & 58.3 & 35.7 & 25.2 & 41.7 \\
FCN-32s RGB-HHA & \textbf{65.3} & \textbf{44.0} & \textbf{33.3} & \textbf{48.6} \\
\bottomrule
\end{tabular}
\label{tab:nyud}
\end{table}

\begin{table}[h!]
\centering
\caption{
Results on SIFT Flow\protect\footnotemark[6] with semantics (center) and geometry (right).
Farabet is a multi-scale convnet trained on class-balanced or natural frequency samples.
Pinheiro is the multi-scale, recurrent convnet {\sc r}CNN$_3$ $(\circ^3)$.
The metric for geometry is pixel accuracy.
}
\begin{tabular}{lccccc}
\toprule
& \pixacc & \classacc & \meanIU & \fwIU & \parbox[b]{0.3in}{geom.\\acc.} \\
\midrule
Liu \etal \cite{sift-flow} & 76.7 & - & - & - & - \\
Tighe \etal \cite{Superparsing} transfer & - & - & - & - & 90.8 \\
Tighe \etal \cite{tighe2013finding} SVM  & 75.6 & 41.1 & - & - & - \\
Tighe \etal  \cite{tighe2013finding} SVM+MRF & 78.6 & 39.2 & - & - & - \\
Farabet \etal \cite{Farabet} natural & 72.3 & 50.8 & - & - & - \\
Farabet \etal \cite{Farabet} balanced & 78.5 & 29.6 & - & - & - \\
Pinheiro \etal \cite{Pinheiro} & 77.7 & 29.8 & - & - & - \\
FCN-8s & \textbf{85.9} & \textbf{53.9} & 41.2 & 77.2 & \textbf{94.6} \\
\bottomrule
\end{tabular}
\label{tab:siftflow}
\end{table}

\begin{table}[h!]
\centering
\caption{
Results on PASCAL-Context for the 59 class task.
\textit{CFM} is convolutional feature masking \cite{CFM} and segment pursuit with the VGG net.
\textit{O$_2$P} is the second order pooling method \cite{O2P} as reported in the \emph{errata} of \cite{mottaghi2014role}.
}
\begin{tabular}{lcccc}
\toprule
59 class  & \pixacc & \classacc & \meanIU & \fwIU \\
\midrule
O$_2$P  & - & - & 18.1 & - \\
CFM     & - & - & 34.4 & - \\
FCN-32s & 65.5 & 49.1 & 36.7 & 50.9 \\
FCN-16s & 66.9 & 51.3 & 38.4 & 52.3 \\
FCN-8s  & \textbf{67.5} & \textbf{52.3} & \textbf{39.1} & \textbf{53.0} \\
\bottomrule
\end{tabular}
\label{tab:pascal-context}
\end{table}

\wordsection{PASCAL-Context} \cite{mottaghi2014role} provides whole scene annotations of PASCAL VOC 2010.
While there are 400+ classes, we follow the 59 class task defined by \cite{mottaghi2014role} that picks the most frequent classes.
We train and evaluate on the training and val sets respectively.
In Table \ref{tab:pascal-context} we compare to the previous best result on this task.
FCN-8s scores 39.1 mean IU for a relative improvement of more than 10\%.

\begin{figure}[h!]
\centering
\setlength{\tabcolsep}{1pt}
\scalebox{0.95}{
\begin{tabular}{cccc}
FCN-8s & SDS \cite{Bharath} & Ground Truth & Image \\
\includegraphics[width=0.12\textwidth]{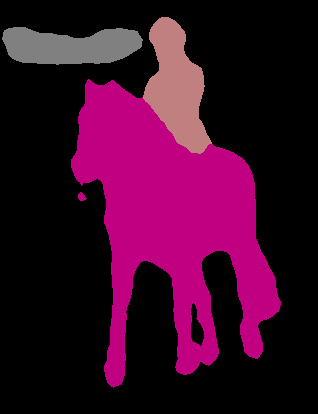} &
\includegraphics[width=0.12\textwidth]{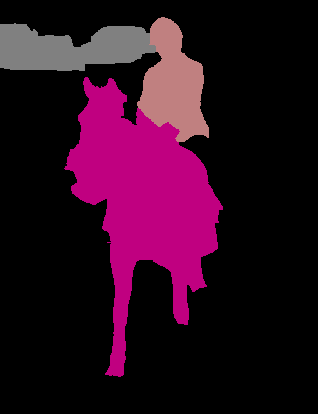} &
\includegraphics[width=0.12\textwidth]{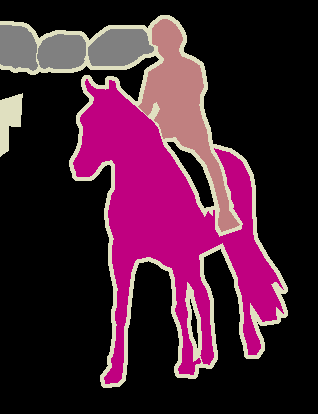} &
\includegraphics[width=0.12\textwidth]{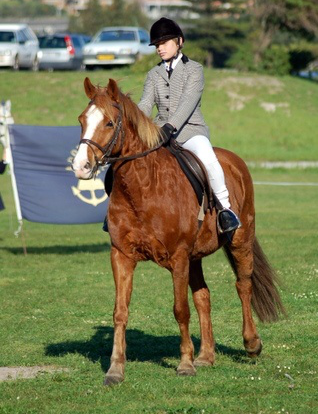} \\
\includegraphics[width=0.12\textwidth]{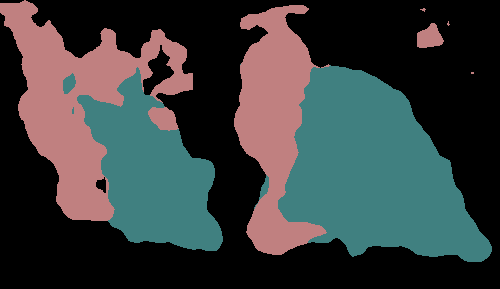} &
\includegraphics[width=0.12\textwidth]{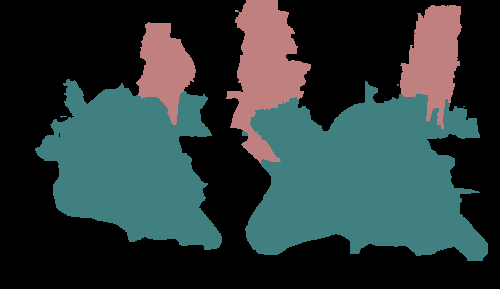} &
\includegraphics[width=0.12\textwidth]{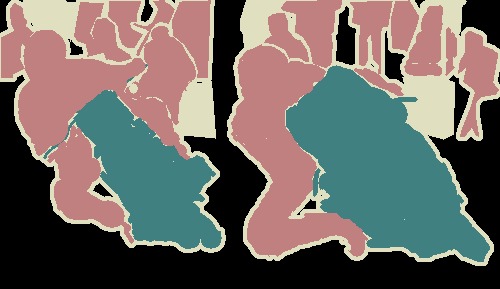} &
\includegraphics[width=0.12\textwidth]{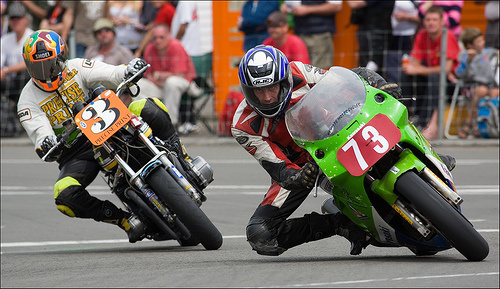} \\
\includegraphics[width=0.12\textwidth]{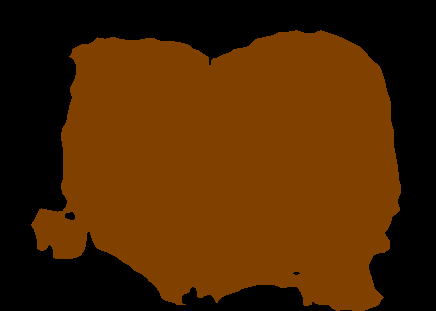} &
\includegraphics[width=0.12\textwidth]{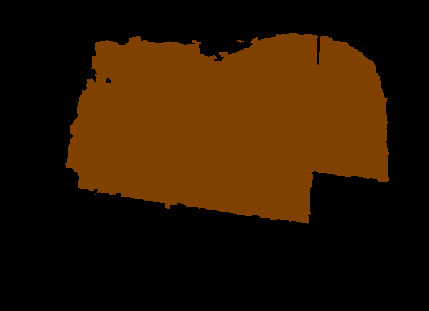} &
\includegraphics[width=0.12\textwidth]{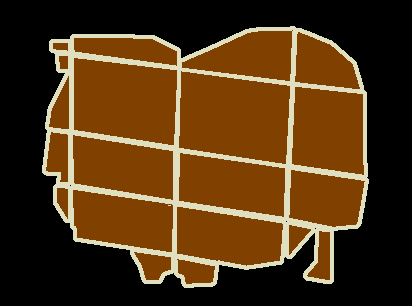} &
\includegraphics[width=0.12\textwidth]{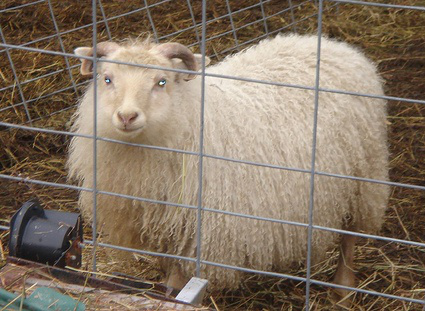} \\
\includegraphics[width=0.12\textwidth]{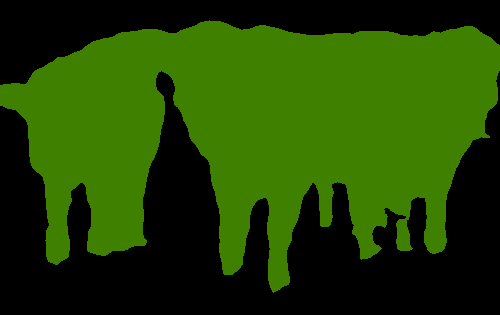} &
\includegraphics[width=0.12\textwidth]{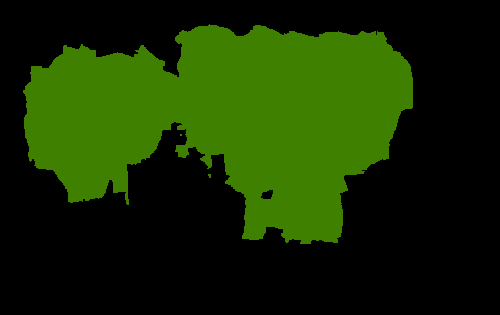} &
\includegraphics[width=0.12\textwidth]{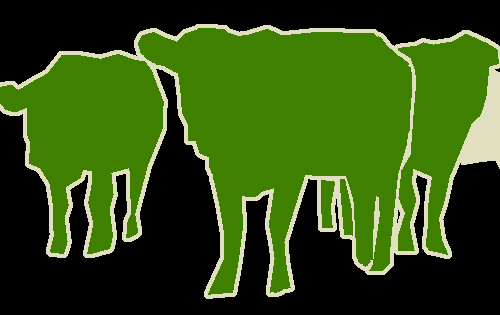} &
\includegraphics[width=0.12\textwidth]{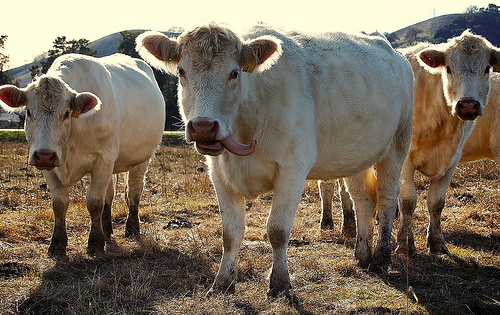} \\
\includegraphics[width=0.12\textwidth]{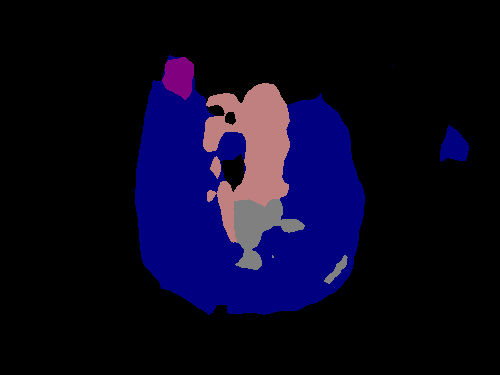} &
\includegraphics[width=0.12\textwidth]{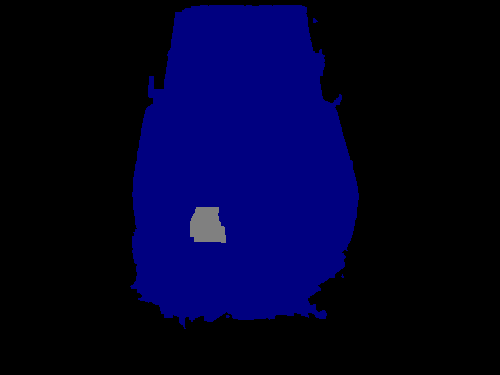} &
\includegraphics[width=0.12\textwidth]{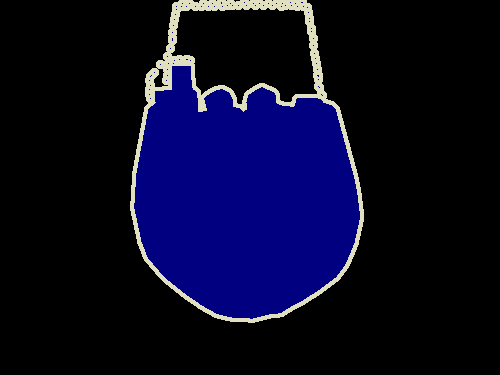} &
\includegraphics[width=0.12\textwidth]{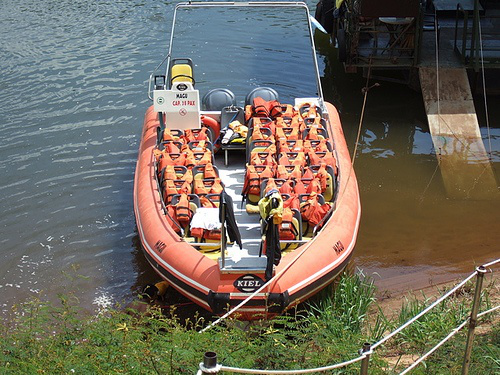} \\
\includegraphics[width=0.12\textwidth]{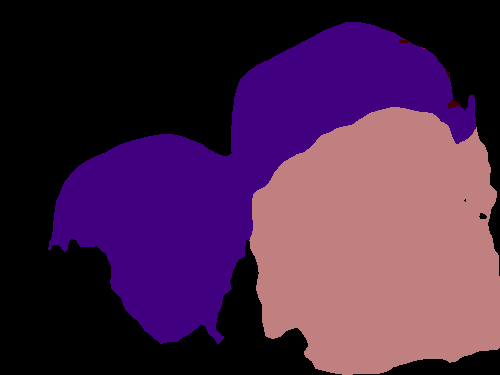} &
\includegraphics[width=0.12\textwidth]{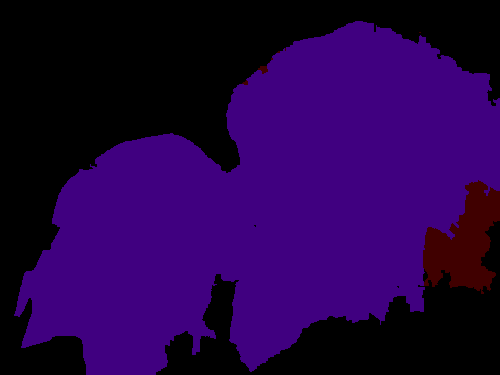} &
\includegraphics[width=0.12\textwidth]{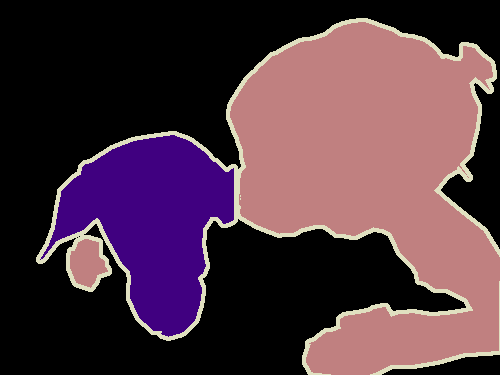} &
\includegraphics[width=0.12\textwidth]{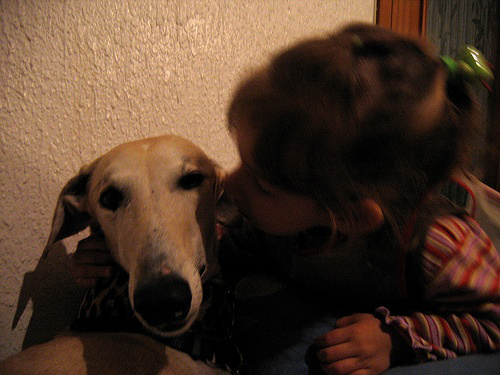} \\
\end{tabular}
}
\caption{
Fully convolutional networks improve performance on PASCAL.
The left column shows the output of our most accurate net, FCN-8s.
The second shows the output of the previous best method by Hariharan \etal \cite{Bharath}.
Notice the fine structures recovered (first row), ability to separate closely interacting objects (second row), and robustness to occluders (third row).
The fifth and sixth rows show failure cases: the net sees lifejackets in a boat as people and confuses human hair with a dog.
}
\end{figure}

\section{Analysis}

We examine the learning and inference of \name s.
Masking experiments investigate the role of context and shape by reducing the input to only foreground, only background, or shape alone.
Defining a ``null'' background model checks the necessity of learning a background classifier for semantic segmentation.
We detail an approximation between momentum and batch size to further tune whole image learning.
Finally, we measure bounds on task accuracy for given output resolutions to show there is still much to improve.

\subsection{Cues}

\begin{table}
\centering
\caption{
  The role of foreground, background, and shape cues.
  All scores are the mean intersection over union metric {\it excluding background}.
  The architecture and optimization are fixed to those of FCN-32s ({\it Reference}) and only input masking differs.
}
\begin{tabular}{lccccc}
\toprule
& \multicolumn{2}{c}{train} & \multicolumn{2}{c}{test} \\
\cmidrule(r){2-3} \cmidrule(l){4-5}
& FG & BG & FG & BG & mean IU \\
\midrule
Reference & keep & keep & keep & keep & 84.8 \\
Reference-FG & keep & keep & keep & mask & 81.0 \\
Reference-BG & keep & keep & mask & keep & 19.8 \\
FG-only & keep & mask & keep & mask & 76.1 \\
BG-only & mask & keep & mask & keep & 37.8 \\
Shape & mask & mask & mask & mask & 29.1 \\
\bottomrule
\end{tabular}
\label{tab:context-iu}
\end{table}

Given the large receptive field size of an \NAME, it is natural to wonder about the relative importance of foreground and background pixels in the prediction.
Is foreground appearance sufficient for inference, or does the context influence the output?
Conversely, can a network learn to recognize a class by its shape and context alone?

\minisection{Masking} To explore these issues we experiment with masked versions of the standard PASCAL VOC segmentation challenge.
We both mask input to networks trained on normal PASCAL, and learn new networks on the masked PASCAL.
See Table \ref{tab:context-iu} for masked results.

Masking the foreground at inference time is catastrophic.
However, masking the foreground during learning yields a network capable of recognizing object segments without observing a single pixel of the labeled class.
Masking the background has little effect overall but does lead to class confusion in certain cases.
When the background is masked during both learning and inference, the network unsurprisingly achieves nearly perfect background accuracy; however certain classes are more confused.
All-in-all this suggests that \NAME s do incorporate context even though decisions are driven by foreground pixels.

\begin{figure}
\makebox[0.242\linewidth]{Image}
\makebox[0.242\linewidth]{Ground Truth}
\makebox[0.242\linewidth]{Output}
\makebox[0.242\linewidth]{Input}

\includegraphics[width=0.5\textwidth]{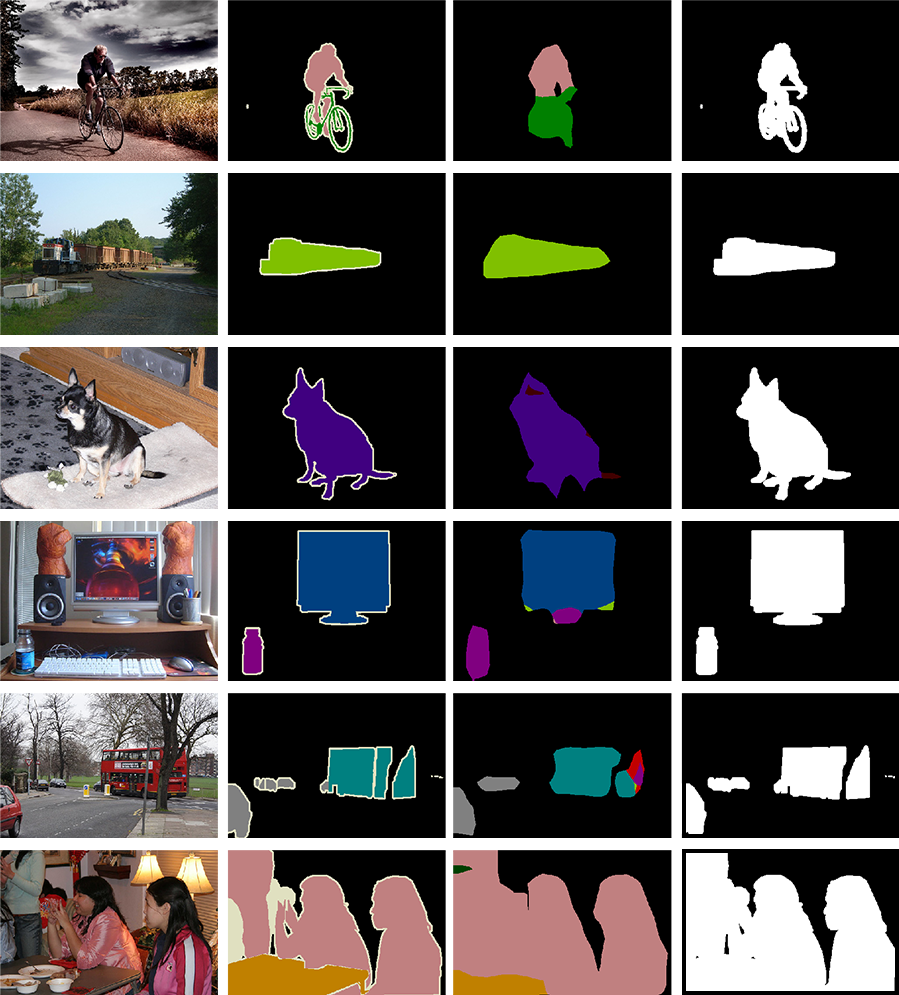}
\caption{
\NAME s learn to recognize by shape when deprived of other input detail.
From left to right: regular image (not seen by network), ground truth, output, mask input.}
\label{fig:shapeseg}
\end{figure}

To separate the contribution of shape, we learn a net restricted to the simple input of foreground/background masks.
The accuracy in this shape-only condition is lower than when only the foreground is masked, suggesting that the net is capable of learning context to boost recognition.
Nonetheless, it is surprisingly accurate.
See Figure \ref{fig:shapeseg}.

\minisection{Background modeling} It is standard in detection and semantic segmentation to have a background model.
This model usually takes the same form as the models for the classes of interest, but is supervised by negative instances.
In our experiments we have followed the same approach, learning parameters to score all classes including background.
Is this actually necessary, or do class models suffice?

To investigate, we define a net with a ``null'' background model that gives a constant score of zero.
Instead of training with the softmax loss, which induces competition by normalizing across classes, we train with the sigmoid cross-entropy loss, which independently normalizes each score.
For inference each pixel is assigned the highest scoring class.
In all other respects the experiment is identical to our FCN-32s on PASCAL VOC.
The null background net scores 1 point lower than the reference FCN-32s and a control FCN-32s trained on all classes including background with the sigmoid cross-entropy loss.
To put this drop in perspective, note that discarding the background model in this way reduces the total number of parameters by less than 0.1\%.
Nonetheless, this result suggests that learning a dedicated background model for semantic segmentation is not vital.

\subsection{Momentum and batch size}
\label{ana:opt}

In comparing optimization schemes for FCNs, we find that ``heavy'' online learning with high momentum trains more accurate models in less wall clock time (see Section \ref{sec:opt}).
Here we detail a relationship between momentum and batch size that motivates heavy learning.

By writing the updates computed by gradient accumulation as a non-recursive sum, we will see that momentum and batch size can be approximately traded off, which suggests alternative training parameters.
Let $g_t$ be the step taken by minibatch SGD with momentum at time $t$,
\[
g_t = - \eta \sum_{i = 0}^{k-1} \nabla_\theta \ell(x_{kt + i}; \theta_{t - 1}) + p g_{t - 1},
\]
where $\ell(x; \theta)$ is the loss for example $x$ and parameters $\theta$, $p < 1$ is the momentum, $k$ is the batch size, and $\eta$ is the learning rate.
Expanding this recurrence as an infinite sum with geometric coefficients, we have
\[
g_t = - \eta \sum_{s = 0}^\infty \sum_{i = 0}^{k-1} p^s \nabla_\theta \ell(x_{k(t - s) + i}; \theta_{t-s}).
\]
In other words, each example is included in the sum with coefficient $p^{\lfloor j / k \rfloor}$, where the index $j$ orders the examples from most recently considered to least recently considered.
Approximating this expression by dropping the floor, we see that learning with momentum $p$ and batch size $k$ appears to be similar to learning with momentum $p'$ and batch size $k'$ if $p^{(1/k)} = p'^{(1/k')}$.
Note that this is not an exact equivalence: a smaller batch size results in more frequent weight updates, and may make more learning progress for the same number of gradient computations.
For typical \NAME\ values of momentum $0.9$ and a batch size of 20 images, an approximately equivalent training regime uses momentum $0.9^{(1/20)} \approx 0.99$ and a batch size of one, resulting in online learning.
In practice, we find that online learning works well and yields better \NAME\ models in less wall clock time.

\subsection{Upper bounds on IU}
\label{sec:ub}

\NAME s achieve good performance on the mean IU segmentation metric even with spatially coarse semantic prediction.
To better understand this metric and the limits of this approach with respect to it, we compute approximate upper bounds on performance with prediction at various resolutions.
We do this by downsampling ground truth images and then upsampling back to simulate the best results obtainable with a particular downsampling factor.
The following table gives the mean IU on a subset\footnotemark[5] of PASCAL 2011 val for various downsampling factors.

\begin{center}
\begin{tabular}{cc}
\toprule
factor & mean IU \\
\midrule
128 & 50.9 \\
64 & 73.3 \\
32 & 86.1 \\
16 & 92.8 \\
8  & 96.4 \\
4  & 98.5 \\
\bottomrule
\end{tabular}
\end{center}

Pixel-perfect prediction is clearly not necessary to achieve mean IU well above state-of-the-art, and, conversely, mean IU is a not a good measure of fine-scale accuracy.
The gaps between oracle and state-of-the-art accuracy at every stride suggest that recognition and not resolution is the bottleneck for this metric.

\section{Conclusion}

\Name s are a rich class of models that address many pixelwise tasks.
\NAME s for semantic segmentation dramatically improve accuracy by transferring pre-trained classifier weights, fusing different layer representations, and learning end-to-end on whole images.
End-to-end, pixel-to-pixel operation simultaneously simplifies and speeds up learning and inference.
All code for this paper is open source in Caffe, and all models are freely available in the Caffe Model Zoo.
Further works have demonstrated the generality of \name s for a variety of image-to-image tasks.

\section*{Acknowledgements}
This work was supported in part by DARPA's MSEE and SMISC programs, NSF awards IIS-1427425, IIS-1212798, IIS-1116411, and the NSF GRFP, Toyota, and the Berkeley Vision and Learning Center.
We gratefully acknowledge NVIDIA for GPU donation.
We thank Bharath Hariharan and Saurabh Gupta for their advice and dataset tools.
We thank Sergio Guadarrama for reproducing GoogLeNet in Caffe.
We thank Jitendra Malik for his helpful comments.
Thanks to Wei Liu for pointing out an issue wth our SIFT Flow mean IU computation and an error in our frequency weighted mean IU formula.

\bibliographystyle{IEEEtran}
\bibliography{bib}

\end{document}